\documentclass[letterpaper, 10 pt, conference]{ieeeconf}  % Comment this line out if you need a4paper

\IEEEoverridecommandlockouts                              % This command is only needed if 
\usepackage{cite}
\usepackage{amsmath,amssymb,amsfonts}
\usepackage{algorithmic}
\usepackage{graphicx}
\usepackage{textcomp}
\def\BibTeX{{\rm B\kern-.05em{\sc i\kern-.025em b}\kern-.08em
    T\kern-.1667em\lower.7ex\hbox{E}\kern-.125emX}}
\usepackage{times}
\usepackage{latexsym}
\usepackage{graphicx}
\usepackage{color}
\usepackage{multirow}
\usepackage{amsmath}
\usepackage{amssymb}
\usepackage[usenames,dvipsnames]{xcolor}
\usepackage[ruled,vlined,linesnumbered]{algorithm2e}
\usepackage{float}
\usepackage{tabularx}
\usepackage{multicol}

\let\labelindent\relax
\usepackage{enumitem}

\newtheorem{definition}{Definition}

\newcommand{\comment}[1]{}

\makeindex

\title{\LARGE \bf
Lazy Modeling of Variants of Token Swapping Problem and Multi-agent Path Finding through Combination of\\Satisfiability Modulo Theories and Conflict-based Search
}

\author{Pavel Surynek$^{1}$ % <-this % stops a space
\thanks{$^{1}$Faculty of Information Technology, Czech Technical University in Prague, Th\'{a}kurova 9, 160 00 Praha 6, Czech Republic
        {\tt\small pavel.surynek@fit.cvut.cz}}%
}

\begin{document}

\maketitle
\thispagestyle{empty}
\pagestyle{empty}

%%%%%%%%%%%%%%%%%%%%%%%%%%%%%%%%%%%%%%%%%%%%%%%%%%%%%%%%%%%%%%%%%%%%%%%%%%%%%%%%
\begin{abstract}

We address item relocation problems in graphs in this paper. We assume items placed in vertices of an undirected graph with at most one item per vertex. Items can be moved across edges while various constraints depending on the type of relocation problem must be satisfied. We introduce a general problem formulation that encompasses known types of item relocation problems such as multi-agent path finding (MAPF) and token swapping (TSWAP). In this formulation we express two new types of relocation problems derived from token swapping that we call token rotation (TROT) and token permutation (TPERM). Our solving approach for item relocation combines satisfiability modulo theory (SMT) with conflict-based search (CBS). We interpret CBS in the SMT framework where we start with the basic model and refine the model with a collision resolution constraint whenever a collision between items occurs in the current solution. The key difference between the standard CBS and our SMT-based modification of CBS (SMT-CBS) is that the standard CBS branches the search to resolve the collision while in SMT-CBS we iteratively add a single disjunctive collision resolution constraint. Experimental evaluation on several benchmarks shows that the SMT-CBS algorithm significantly outperforms the standard CBS. We also compared SMT-CBS with a modification of the SAT-based MDD-SAT solver that uses an eager modeling of item relocation in which all potential collisions are eliminated by constrains in advance. Experiments show that lazy approach in SMT-CBS produce fewer constraint than MDD-SAT and also achieves faster solving run-times.

\end{abstract}

%%%%%%%%%%%%%%%%%%%%%%%%%%%%%%%%%%%%%%%%%%%%%%%%%%%%%%%%%%%%%%%%%%%%%%%%%%%%%%%%
\section{INTRODUCTION AND MOTIVATION}

item relocation problems in graphs such as {\em token swapping} (TSWAP) \cite{DBLP:conf/walcom/KawaharaSY17,DBLP:conf/stacs/BonnetMR17}, {\em multi-agent path finding} (MAPF) \cite{Ryan07,standley2010finding,DBLP:conf/icra/YuL13}, or pebble motion on graphs (PMG) \cite{WILSON197486,DBLP:conf/focs/KornhauserMS84} represent important combinatorial problems in artificial intelligence with specific applications in robots. We assume that multiple distinguishable items are placed in vertices of an undirected graph such that at most one item is placed in each vertex. Items can be moved between vertices across edges while problem specific rules are observed during movement. For instance, PMG and MAPF usually requires that items (pebbles/agents) are moved to unoccupied neighbors only. TSWAP on the other hand permits only swaps of pairs of tokens along edges while more complex movements are forbidden. The task in item relocation problems is to reach a given goal configuration of items from a given starting configuration using allowed movements.

In this paper we focus on optimal solving of item relocation problems with respect to cummulative objective functions. Two cumulative objective functions are commonly used in MAPF and TSWAP - {\em sum-of-costs} \cite{ICTSJUR,DBLP:conf/esa/MiltzowNORTU16} and {\em makespan} \cite{DBLP:conf/ictai/Surynek14}. Sum-of-costs corresponds to the total cost of all movements performed until the goal configuration in reached - traversal of an edge by an item has unit cost typically. Makespan calculates the total number of time-steps until the goal is reached. In both cases we trying to minimize the objective which in the case of sum-of-costs intuitively corresponds to energy minimization while the minimization of makespan corresponds to minimization of time.

Many practical problems from robotics can be interpreted as an item relocation problem. Examples include discrete multi-robot navigation and coordination \cite{DBLP:conf/iros/LunaB10}, item rearrangement in automated warehouses \cite{DBLP:journals/tase/BasileCC12}, ship collision avoidance \cite{ShipAviodance2014}, or formation maintenance and maneuvering of aerial vehicles \cite{DBLP:conf/icra/ZhouS15}.

\subsection{Contrubutions}

The contribution of this paper consists in suggesting a general framework for defining and solving item relocation problems based on {\em satisfiability modulo theories} (SMT) \cite{DBLP:journals/constraints/BofillPSV12} and {\em conflict-based search} (CBS) \cite{SharonSFS15}. We used the framework to define two problems derived from TSWAP: {\em token rotation} (TROT) and {\em token permutation} (TPERM) where instead of swapping tokens rotations along non-trivial cycles and arbitrary permutations of tokens are permitted. A novel algorithm called SMT-CBS that combines ideas from CBS and SMT is suggested and experimentally evaluated. Tests on standard synthetic benchmarks indicate that SMT-CBS outperforms previous state-of-the-art SAT-based algorithm for optimal MAPF solving MDD-SAT \cite{SurynekFSB16,surynek2016empirical}.

The organization of the paper is as follows. We first introduce TSWAP and MAPF problems formally. Then prerequisities for constructing SMT framework for item relocation problems are recalled, the CBS algorithm and the MDD-SAT algorithm. On top of this, the combination of CBS and MDD-SAT is developed resulting in the SMT-CBS algorithm. Finally experimental evaluation of all concerned algorithms CBS, MDD-SAT, and SMT-CBS is presented.

\section{BACKGROUND}

We recall {\em multi-agent path finding} and {\em token swapping} as they appear in the literature in this section.

{\em Multi-agent path finding} (MAPF) problem \cite{DBLP:conf/aiide/Silver05,DBLP:journals/jair/Ryan08} consists of an undirected graph $G=(V,E)$ and a set of agents $A=\{a_1, a_2, ..., a_k\}$ such that $|A|<|V|$. Each agent is placed in a vertex so that at most one agent resides in each vertex. The placement of agents is denoted $\alpha: A \rightarrow V$.  Next we are given nitial configuration of agents $\alpha_0$ and goal configuration $\alpha_+$.

\begin{figure}[h]
    \centering
    \includegraphics[trim={5.2cm 23cm 4.8cm 3.5cm},clip,width=0.55\textwidth]{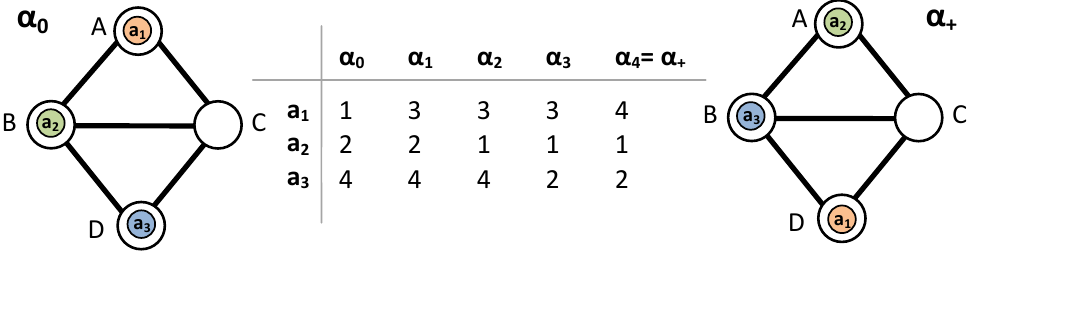}
    \vspace{-1.0cm}\caption{A MAPF instance with three agents $a_1$, $a_2$, and $a_3$.}
    \label{figure-MAPF}
\end{figure}

At each time step an agent can either {\em move} to an adjacent location or {\em wait} in its current location. The task is to find a sequence of move/wait actions for each agent $a_i$, moving it from $\alpha_0(a_i)$ to $\alpha_+(a_i)$ such that agents do not {\em conflict}, i.e., do not occupy the same location at the same time. Typically, an agent can move into adjacent unoccupied vertex provided no other agent enters the same target vertex but other rules for movements are used as well. An example of MAPF instance is shown in Figure \ref{figure-MAPF}.

The following definition formalizes the commonly used {\em move-to-unoccupied} movement rule in MAPF.

\begin{definition}
    {\bf Movement in MAPF.}
    Configuration $\alpha'$ results from $\alpha$ if and only if the following conditions hold: (i) $\alpha(a) = \alpha'(a)$ or $\{\alpha(a),\alpha'(a)\} \in E$ for all $a \in A$ (agents wait or move along edges); (ii) for all $a \in A$ it holds that if ${\alpha(a) \neq \alpha'(a)} \Rightarrow {\alpha'(a) \neq \alpha(a')}$ for all $a' \in A$ (target vertex must be empty); and (iii) for all $a,a' \in A$ it holds that if ${a \neq a'} \Rightarrow {\alpha'(a) \neq \alpha'(a')}$ (no two agents enter the same target vertex).
    \label{def:movement}
\end{definition}

Solving the MAPF instance is to search for a sequence of configurations $[\alpha_0,\alpha_1,...,\alpha_{\mu}]$ such that  $\alpha_{i+1}$ results using valid movements from $\alpha_{i}$ for $i=1,2,...,\mu-1$, and $\alpha_{\mu}=\alpha_+$. 

In many aspects, {\em token swapping problem (TSWAP)} (also known as {\em sorting on graphs}) \cite{DBLP:conf/fun/YamanakaDIKKOSSUU14} is similar to MAPF. It represents a generalization of sorting problems \cite{DBLP:journals/jal/Thorup02}. While in the classical sorting problem we need to obtain linearly ordered sequence of elements by swapping any pair of elements, in the TSWAP problem we are allowed to swap elements at selected pairs of positions only.

\begin{figure}[t]
    \centering
    \includegraphics[trim={5.2cm 23cm 5.2cm 3.7cm},clip,width=0.55\textwidth]{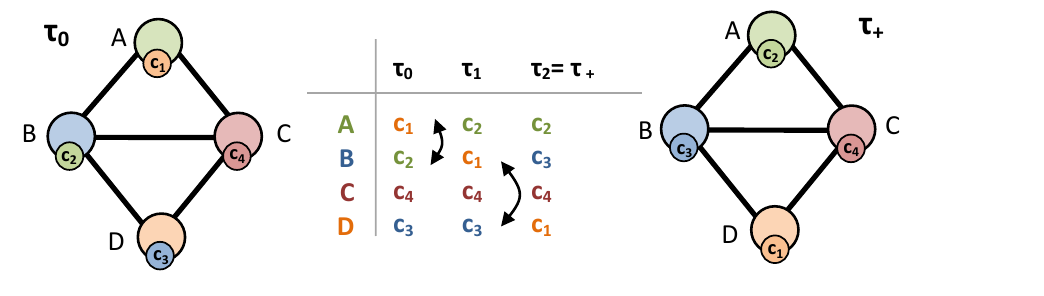}
    \vspace{-0.8cm}\caption{A TSWAP instance. A solution consisting of two swaps is shown.}
    \label{figure-TSWAP}
\end{figure}

Using a modified notation from \cite{DBLP:journals/tcs/YamanakaDIKKOSS15} the TSWAP each vertex in $G$ is assigned a color in $C = \{c_1,c_2,...,c_h\}$ via $\tau_+:V\rightarrow C$. A token of a color in $C$ is placed in each vertex. The task is to transform a current token placement into the one such that colors of tokens and respective vertices of their placement agree. Desirable token placement can be obtained by swapping tokens on adjacent vertices in $G$. See Figure \ref{figure-TSWAP} for an example instance of TSWAP.

We denote by $\tau:V\rightarrow C$ colors of tokens placed in vertices of $G$. That is, $\tau(v)$ for $v \in V$ is a color of a token placed in $v$. Starting placement of tokens is denoted as $\tau_0$; the goal token placement corresponds to $\tau_+$. Transformation of one placement to another is captured by the concept of {\em adjacency} defined as follows \cite{DBLP:journals/tcs/YamanakaDIKKOSS15,DBLP:conf/walcom/YamanakaDHKNOSS17}:

\begin{definition}
     {\bf adjacency in TSWAP}
    Token placements $\tau$ and $\tau'$ are said to be adjacent if there exists a subset of non-adjacent edges $F \subseteq E$  such that $\tau(v) =  \tau'(u)$ and $\tau(u) = \tau'(v)$ for each $\{u,v\} \in F$ and for all other vertices $w \in V \setminus \bigcup_{\{u,v\} \in F}{\{u,v\}}$ it holds that $\tau(w) = \tau'(w)$. \footnote{The presented version of adjacency is sometimes called {\em parallel} while a term adjacency is reserved for the case with $|F|=1$.}
    \label{def:adjacency-tswap}
\end{definition}

The task in TSWAP is to find a swapping sequence of token placements $[\tau_0, \tau_1, ..., \tau_m]$ such that $\tau_m = \tau_+$ and $\tau_i$ and $\tau_{i+1}$ are adjacent for all $i=0,1,..., m-1$. It has been shown that for any initial and goal placement of tokens $\tau_0$ and $\tau_+$ respectively there is a swapping sequence transforming $\tau_0$ and $\tau_+$ containing $\mathcal{O}({|V|}^2)$ swaps \cite{YamanakaComplex2016}. The proof is based on swapping tokens on a spanning tree of $G$. Let us note that the above bound is tight as there are instances consuming $\Omega({|V|}^2)$ swaps. It is also known that finding swapping sequence that has as few swaps as possible is an NP-hard problem.

If each token has a different color we do not distinguish between tokens and theer colors $c_i$; that is, we will refer to a token $c_i$.

Observe, that the operational meaning of agents and tokens in MAPF and TSWAP is similar. They both occupy vertices of the graph and no two of them can share a vertex.  Hence works studying relation of both problems from the practical solving perspective have appeared recently \cite{DBLP:conf/ictai/Surynek18}.

\section{RELATED WORK}

Although many works sudying TSWAP from the theoretical point of view exist \cite{YamanakaComplex2016,DBLP:conf/esa/MiltzowNORTU16,DBLP:conf/stacs/BonnetMR17} practical solving of the problem started only lately.
In \cite{DBLP:conf/ictai/Surynek18} optimal solving of TSWAP by adapted algorithms from MAPF has been suggested. Namely {\em conflict-based search} (CBS) \cite{DBLP:conf/aaai/SharonSFS12,SharonSFS15} and {\em propositional satisfiability-based} (SAT) \cite{Biere:2009:HSV:1550723} MDD-SAT \cite{SurynekFSB16,surynek2016empirical} originally developed for MAPF have been modified for TSWAP.

\subsection{Search for Optimal Solutions}
We will commonly use the {\em sum-of-costs} objective funtion in all problems studied in this paper. The following definition introduces the sum-of-costs objective in MAPF. However, analogical definition can be introduced for TSWAP too.

\begin{definition}
	{\bf Sum-of-costs} (denoted $\xi$) is the summation, over all
	agents, of the number of time steps required to reach the goal
	vertex~\cite{dresner2008aMultiagent,standley2010finding,DBLP:journals/ai/SharonSGF13,CBSJUR}.
	Formally, $\xi = \sum_{i=1}^k{\xi(path(a_i))}$, where $\xi(path(a_i))$ is an
	\textit{individual path cost} of agent $a_i$ connecting $\alpha_0(a_i)$ calculated as the number of edge traversals and wait actions.			\footnote{The notation $path(a_i)$ refers to path in the form of a seqeunce of vertices and edges connecting $\alpha_0(a_i)$ and $		\alpha_	+(a_i)$ while $\xi$ assigns the cost to a given path.}
\end{definition}

Observe that in the sum-of-costs we accumulate the cost of wait actions for items not yet reaching their goal vertices. Also observe that one swap in the TSWAP problem correspond to the cost of 2 as two tokens traverses single edge. Let us note that all algorithms and concepts we use can be modified for different cummulative objective functions like makespan or the total number of moves/swaps etc.

Feasible solution of a solvable MAPF instance can be found in polynomial time \cite{WILSON197486,DBLP:conf/focs/KornhauserMS84}; precisely the worst case time complexity of most practical algorithms for finding feasible solutions is $\mathcal{O}({|V|}^3)$ (asymptotic size of the solution is also $\mathcal{O}({|V|}^3)$) \cite{DBLP:conf/icra/Surynek09,DBLP:conf/ictai/Surynek09,DBLP:journals/ci/Surynek14,luna2011efficient,LunaB11,DBLP:journals/jair/WildeMW14}. This is also asymtotically best what can be done as there are MAPF instances requiring $\Omega({|V|}^2)$ moves. As with TSWAP, finding optimal MAPF solutions with respect to various cummulative objectives is NP-hard \cite{DBLP:conf/aaai/RatnerW86,DBLP:conf/aaai/Surynek10,DBLP:journals/corr/YuL15c}.

\subsection{Conflict-based Search}

CBS uses the idea of resolving conflicts lazily; that is, a solution of MAPF instance is not searched against the complete set of movement constraints that forbids collisions between agents but with respect to initially empty set of collision forbidding constraints that gradually grows as new conflicts appear. The advantage of CBS is that it can find a valid solution before all constraints are added.

The high level of CBS searches a {\em constraint tree} (CT) using a priority queue in breadth first manner. CT is a binary tree where each node $N$ contains a set of collision avoidance constraints $N.constraints$ - a set of triples $(a_i,v,t)$ forbidding occurrence of agent $a_i$ in vertex $v$ at time step $t$, a solution $N.paths$ - a set of $k$ paths for individual agents, and the total cost $N.\xi$ of the current solution.

The low level process in CBS associated with node $N$ searches paths for individual agents with respect to set of constraints $N.constraints$. For a given agent $a_i$, this is a standard single source shortest path search from $\alpha_0(a_i)$ to $\alpha_+(a_i)$ that avoids a set of vertices $\{v \in V|(a_i,v,t) \in N.constraints\}$ whenever working at time step $t$. For details see \cite{SharonSFS15}.

CBS stores nodes of CT into priority queue $\textsc{Open}$ sorted according to ascending costs of solutions. At each step CBS takes node $N$ with lowest cost from $\textsc{Open}$ and checks if $N.paths$ represents paths that are valid with respect to movements rules in MAPF. That is, if there are any collisions between agents in $N.paths$. If there is no collision, the algorithms returns valid MAPF solution $N.paths$. Otherwise the search branches by creating a new pair of nodes in CT - successors of $N$. Assume that a collision occurred between agents $a_i$ and $a_j$ in vertex $v$ at time step $t$. This collision can be avoided if either agent $a_i$ or agent $a_j$ does not reside in $v$ at timestep $t$. These two options correspond to new successor nodes of $N$ - $N_1$ and $N_2$ that inherits set of conflicts from $N$ as follows: $N_1.conflicts = N.conflicts \cup \{(a_i,v,t)\}$ and $N_2.conflicts = N.conflicts \cup \{(a_j,v,t)\}$. $N_1.paths$ and $N_1.paths$ inherit path from $N.paths$ except those for agent $a_i$ and $a_j$ respectively. Paths for $a_i$ and $a_j$ are recalculated with respect to extended sets of conflicts $N_1.conflicts$ and $N_2.conflicts$ respectively and new costs for both agents $N_1.\xi$ and $N_2.\xi$ are determined. After this $N_1$ and $N_2$ are inserted into the priority queue $\textsc{Open}$.

The pseudo-code of CBS is listed as Algorithm \ref{alg-CBS}. One of crucial steps occurs at line 16 where a new path for colliding agents $a_i$ and $a_j$ is constructed with respect to an extended set of conflicts. Notation $N.paths(a)$ refers to the path of agent $a$.

\begin{algorithm}[t!]
\begin{footnotesize}
\SetKwBlock{NRICL}{CBS ($G=(V,E),A,\alpha_0,\alpha_+)$}{end} \NRICL{
    $R.constraints \gets \emptyset$ \\
    $R.paths \gets$ $\{$shortest path from $\alpha_0(a_i)$ to $\alpha_+(a_i) | i = 1,2,...,k\}$\\
    $R.\xi \gets \sum_{i=1}^k{\xi(N.paths(a_i))}$ \\
    insert $R$ into $\textsc{Open}$ \\
    \While {$\textsc{Open} \neq \emptyset$} {
        $N \gets$ min($\textsc{Open}$)\\
        remove-Min($\textsc{Open}$)\\
        $collisions \gets$ validate($N.paths$)\\
        \If {$collisions = \emptyset$}{
            \Return $N.paths$\\
        }
        let $(a_i,a_j,v,t) \in collisions$\\
        
        \For {each $a \in \{a_i,a_j\}$}{
       	$N'.constraints \gets N.constraints \cup \{(a,v,t)\}$\\
        	$N'.paths \gets N.paths$\\
        	update($a$, $N'.paths$, $N'.conflicts$)\\
		$N'.\xi \gets \sum_{i=1}^k\xi{(N'.paths(a_i))}$\\
		insert $N'$ into $\textsc{Open}$ \\
        }
     }
} \caption{Basic CBS algorithm for MAPF solving} \label{alg-CBS}
\end{footnotesize}
\end{algorithm}

The CBS algorithm ensures finding sum-of-costs optimal solution. Detailed proofs of this claim can be found in \cite{CBSJUR}.

\subsection{SAT-based Approach}

An alternative approach to optimal MAPF solving as well as to TSWAP solving is represented by reduction of MAPF to propositional satisfiability (SAT) \cite{DBLP:conf/pricai/Surynek12,DBLP:conf/ictai/Surynek12}. The idea is to construct a propositional formula such $\mathcal{F(\xi)}$ such that it is satisfiable if and only if a solution of a given MAPF of sum-of-costs $\xi$ exists. Moreover, the approach is constructive; that is, $\mathcal{F(\xi)}$ exactly reflects the MAPF instance and if satisfiable, solution of MAPF can be reconstructed from satisfying assignment of the formula.

Being able to construct such formula $\mathcal{F}$ one can obtain optimal MAPF solution by checking satisfiability of $\mathcal{F}(0)$, $\mathcal{F}(1)$, $\mathcal{F}(2)$,... until the first satisfiable $\mathcal{F(\xi)}$ is met. This is possible due to monotonicity of MAPF solvability with respect to increasing values of common cummulative objectives such as the sum-of-costs. In practice it is however impractical to start at 0; lower bound estimation is used instead - sum of lengths of shortest paths can be used in the case of sum-of-costs. The framework of SAT-based solving is shown in pseudo-code in Algorithm \ref{alg-SAT}.

\begin{algorithm}[t]
\begin{footnotesize}
\SetKwBlock{NRICL}{CBS ($G=(V,E),A,\alpha_0,\alpha_+)$}{end} \NRICL{
    $paths \gets$ $\{$shortest path from $\alpha_0(a_i)$ to $\alpha_+(a_i) | i = 1,2,...,k\}$ \\
    $\xi \gets \sum_{i=1}^k{\xi(N.paths(a_i))}$ \\
    \While {$True$}{
        $\mathcal{F}(\xi) \gets$ encode$(\xi,G,A,\alpha_0, \alpha_+)$\\
        $assignment \gets$ consult-SAT-Solver$(\mathcal{F}(\xi))$\\
        \If {$assignment \neq$ UNSAT}{
        	$paths \gets$ extract-Solution$(assignment)$\\
        	\Return $paths$\\
        }
        $\xi \gets \xi + 1$\\
    }
} \caption{Framework of SAT-based MAPF solving} \label{alg-SAT}
\end{footnotesize}
\end{algorithm}

The advantage of the SAT-based approach is that state-of-the-art SAT solvers can be used for determinig satisfiability of $\mathcal{F}(\xi)$ \cite{DBLP:conf/sat/AudemardLS13} and any progress in SAT solving hence can be utilized for increasing efficiency of MAPF solving.

\subsection{Multi-value Decision Diagrams - MDD-SAT}

Construction of $\mathcal{F(\xi)}$ relies on time expansion of underlying graph $G$ \cite{DBLP:journals/amai/Surynek17}. Having $\xi$, the basic variant of time expansion determines the maximum number of time steps $\mu$ (also refered to as a {\em makespan}) such that every possible solution of the given MAPF with the sum-of-costs less than or equal to $\xi$ fits within $\mu$ timestep (that is, no agent is outside its goal vertex after $\mu$ timestep if the sum-of-costs $\xi$ is not to be exceeded).

Time expansion itself makes copies of vertices $V$ for each timestep $t=0,1,2,...,\mu$. That is, we have vertices $v^t$ for each $v \in V$ time step $t$. Edges from $G$ are converted to directed edges interconnecting timesteps in time expansion. Directed edges $(u^t,v^{t+1})$ are introduced for $t=1,2,...,\mu-1$ whenever there is $\{u,v\} \in E$. Wait actions are modeled by introducing edges $(u^t,t^{t+1})$. A directed path in time expansion corresponds to trajectory of an agent in time. Hence the modeling task now consists in construction of a formula in which satisfying assignments correspond to directed paths from $\alpha_0^0(a_i)$ to $\alpha_+^\mu(a_i)$ in time expansion.

Assume that we have time expansion $TEG_i=(V_i,E_i)$ for agent $a_i$. Propositional variable $\mathcal{X}_v^t(a_j)$ is introduced for every vertex $v^t$ in $V_i$. The semantics of $\mathcal{X}_v^t(a_i)$ is that it is $True$ if and only if agent $a_i$ resides in $v$ at time step $t$. Similarly we introduce $\mathcal{E}_u,v^t(a_i)$ for every directed edge $(u^t,v^{t+1})$ in $E_i$. Analogically the meaning of $\mathcal{E}_{u,v}^t(a_i)$ is that is $True$ if and only if agent $a_i$ traverses edge $\{u,v\}$ between time steps $t$ and $t+1$.

Finally constraints are added so that truth assignment are restricted to those that correspond to valid solutions of a given MAPF. The detailed list of constraints is given in \cite{SurynekFSB16}. We here just illustrate the modeling by showing few representative constraints. For example there is a constraint stating that if agent $a_i$ appears in vertex $u$ at time step $t$ then it has to leave through exactly one edge $(u^t,v^{t+1})$. This can be established by following constraints:

\begin{equation}
   {  \mathcal{X}_u^t(a_i) \Rightarrow \bigvee_{(u^t,v^{t+1}) \in E_i}{\mathcal{E}^t_{u,v}(a_i),}
   }
   \label{eq:basic-start}
\end{equation}
\begin{equation}
   {  \sum_{v^{t+1}|(u^t,v^{t+1}) \in E_i }{\mathcal{E}_{u,v}^t{(a_i)} \leq 1}
   }
   \label{eq-1}
\end{equation}

Similarly, the target vertex of any movement except wait action must be
empty. This is ensured by the following constraint for every $(u^t,v^{t+1}) \in E_i$:
\begin{equation}
   {  \mathcal{E}_{u,v}^t(a_i) \Rightarrow \bigwedge_{a_j \in A \wedge a_j \neq a_i \wedge v^{t} \in V_j}{\neg \mathcal{X}_v^{t}(a_j)}
   }
   \label{eq-2}
\end{equation}
\vspace{0.25cm}

Other constraints ensure that truth assignments to variables per individual agents form paths. That is if agent $a_i$ enters an edge it must leave the edge at the next time step.

\begin{equation}
   {  \mathcal{E}^t_{u,v}(a_i) \Rightarrow \mathcal{X}^t_v(a_i) \wedge \mathcal{X}^{t+1}_v(a_i)
   }
   \label{eq-3}
\end{equation}
\vspace{0.25cm}

Agents do not collide with each other; the following constraint is introduced for every $v \in V$ and timestep $t$:

\begin{equation}
   {  \sum_{i=1,2,...,k |v^t \in V_i}{\mathcal{X}^t_v(a_i)}
    }
    \label{eq-4}
\end{equation}
\vspace{0.25cm}

A common measure how to reduce the number of decision variables derived from the time expansion is the use of {\em multi-value decision diagrams} (MDDs) \cite{DBLP:journals/ai/SharonSGF13}. The basic observation that holds for MAPF and other item relocation problems is that a token/agent can reach vertices in the distance $d$ (distance of a vertex is measured as the length of the shortest path) from the current position of the agent/token no earlier than in the $d$-th time step. Analogical observation can be made with respect to the distance from the goal position.

Above observations can be utilized when making the time expansion of $G$. For a given agent or token, we do not need to consider all vertices at time step $t$ but only those that are reachable in $t$ timesteps from the initial position and that ensure that the goal can be reached in the remaining $\sigma - t$ timesteps. This idea can reduce the size the expansion graph significantly and consequently can reduce the size of the Boolean formula by eliminating $\mathcal{X}(a)_v^t$ and $\mathcal{E}(a)_{u,v}^t$ variables correspoding to unreachable vertices $u$ and $v$.

\begin{figure}[h]
    \centering
    \vspace{-0.2cm}
    \includegraphics[trim={4.2cm 22.7cm 5.1cm 3cm},clip,width=0.48\textwidth]{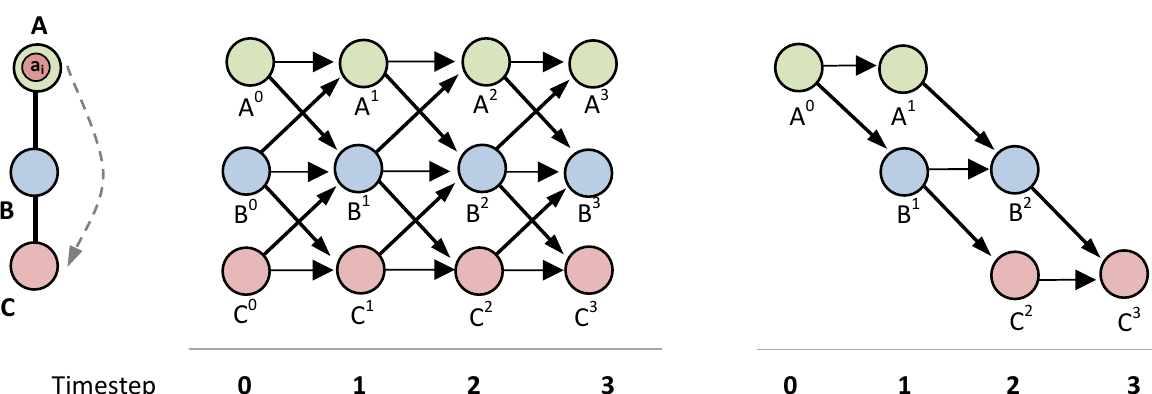}
    \vspace{-0.2cm}
    \caption{An example of time expansion and MDD expansion for agent $a_1$.}
    \label{fig-MDD}
\end{figure}

A comparison of standard time expansion and MDD expansion in MAPF for agent ($a_i$) is shown in Figure \ref{fig-MDD}.

The combination of SAT-based approach and MDD time expansion led to the MDD-SAT algorithm described in \cite{SurynekFSB16} that currently represent state-of-the-art in SAT-based MAPF solving.

\section{GENERALIZATIONS OF ITEM RELOCATION}

Although the differences between MAPF and TSWAP led to different worst case time complexities in algorithms for finding feasible solutions, problems differ only in local understanding of conflicts reflected in different movement rules in fact. This immediately inspired us to suggest various modifications of movement rules.

We define two problems derived from MAPF and TSWAP: {\em token rotation} (TROT) and {\em token permutation} (TPERM) \footnote{These problems have been considered in the literature in different contexts already (for example in \cite{DBLP:conf/wafr/YuR14}). But not from the practical solving perspective in focused on finding optimal solutions.}.

\subsection{Token Rotation and Token Permutation}

A swap of pair of tokens can be interpreted as a rotation along a trivial cycle consisting of single edge. We can generalize this towards longer cycles. The TROT problem permits rotations along longer cycles but forbids trivial cycles; that is, rotations along triples, quadruples, ... of vertices is allowed but swap along edges are forbidden.

\begin{definition}
    {\bf Adjacency in TROT.}
    Token placements $\tau$ and $\tau'$ are said to be adjacent in TROT if there exists a subset of edges $F \subseteq E$ such that 
    components $C_1,C_2,...,C_p$ of induced sub-graph $G[F]$ satisfy following conditions:
     \begin{enumerate}[label=(\roman*)]
     \item {$C_j=(V_j^C,E_j^C)$ such that $V_j^C={w^j_1,w^j_2, ...,w^j_{n_j}}$ with $n_j \leq 3$ and\\ $E_j^C=\{\{w^j_1,w^j_2\};\{w^j_2,w^j_3\};...; \{w^j_{n_j},w^j_1\}\}$
     \\(components are cycles of length at least 3)}
     \item {$\tau(w^j_1) =  \tau'(w^j_2)$, $\tau(w^j_3) =  \tau'(w^j_3)$, ..., $\tau(w^j_{n_j}) =  \tau'(w^j_1)$
     \\(colors are rotated in the cycle one position forward/backward)}
    \end{enumerate}
    \label{def:adjacency-trot}
\end{definition}

The rest of the definition of a TROT instance is analogical to TSWAP.

Similarly we can define TPERM by permitting all lengths of cycles. The formal definition of {\em adjacency} in TPERM is almost the same as in TROT except relaxing the constraint on cycle lenght, $n_j \leq 2$.

We omit here complexity considerations for TROT and TPERM for the sake of brevity. Again it holds that a feasible solution can be found in polynomial time but the optimal cases remain intractable in general.

Both approaches - SAT-based MDD-SAT as well as CBS - can be adapted for solving TROT and TPERM without modifying their top level design. Only local modification of how movement rules of each problem are reflected in algorithms is necessary. In case of CBS, we need to define what does it mean a conflict in TROT and TPERM. In MDD-SAT different movement constraints can be encoded directly.

Motivation for studying these item relocation problems is the same as for MAPF. In many real-life scenarios it happens that items or agents enters positions being simultaneously vacated by other items (for example mobile robots often ). This is exactly the property captured formally in above definitions.

\subsection{Adapting CBS and MDD-SAT}

Both CBS and MDD-SAT can be modified for optimal solving of TSWAP, TROT, and TPERM (with respect to sum-of-costs but other cumulative objectives are possible as well). Different movement rules can be reflected in CBS and MDD-SAT algorithms without modifying their high level framework.

\subsubsection{Different Conflicts in CBS}

In CBS, we need to modify the understanding of conflict between agents/tokens. Proofs of soundness of these changes are omitted.

\noindent
{\bf TPERM:} The easiest case is TPERM as it is least restrictive. We merely forbid simultaneous occurrence of multiple tokens in a vertex - this situation is understood as a collision in TPERM and conflicts are derived from it. If a collision $(c_i,c_j,v,t)$ between tokens $c_i$ and $c_j$ occurs in $v$ at time step $t$ then we introduce conflicts $(c_i,v,t)$ and $(c_j,v,t)$ for $c_i$ and $c_j$ respectively. \footnote{Formally this is the same as in MAPF, but in addition to this MAPF checks vacancy of the target vertex which may cause more colliding situations.}

\noindent
{\bf TSWAP:} This problem takes conflicts from TPERM but adds new conflicts that arise from doing something else than swapping \cite{DBLP:conf/ictai/Surynek18}. Each time edge $\{u,v\}$ is being traversed by token $c_i$ between time steps $t$ and $t+1$, a token residing in $v$ at time step $t$, that is $\tau_t(v)$, must go in the opposite direction from $v$ to $u$. If this is not the case, then a so called {\em edge collision} involving edge $\{u,v\}$ occurs and corresponding {\em edge conflicts} $(c_i,(u,v),t)$ and $(\tau_t(v),(v,u),t)$ are introduced for agents $c_i$ and $\tau_t(v)$ respectively.

Edge conflicts must be treated at the low level of CBS. Hence in addition to forbidden vertices at given time-steps we have forbidden edges between given time-steps.

\noindent
{\bf TROT:} The treatment of conflicts will be complementary to TSWAP in TROT. Each time edge $\{u,v\}$ is being traversed by token $c_i$ between time steps $t$ and $t+1$, a token residing in $v$ at time step $t$, that is $\tau_t(v)$, must go anywhere else but not to $u$. If this is not the case, then we again have edge collision $(c_i,\tau_t(v),$\{u,v\}$,t)$ which is treated in the same way as above.

\subsubsection{Encoding Changes in MDD-SAT}

In MDD-SAT, we need to modify encoding of movement rules in the propositional formula  $\mathcal{F(\xi)}$. Again, proofs of soundness of the following changes are omitted.

\noindent
{\bf TPERM:} This is the easiest case for MDD-SAT too. We merely remove all constrains requiring tokens to move into vacant vertices only. That is we remove clauses (\ref{eq-2}).

\noindent
{\bf TSWAP:} It inherits changes from TPERM but in addition to that we need to carry out swaps properly. For this edge variables
$\mathcal{E}^t_{u,v}(c_i)$ will be utilized. Following constraint will be introduced for every $\{u^t,v^{t+1}\} \in E_i$ (intuitively, if token $c_i$ traverses $\{u,v\}$ some other token $c_j$ traverses $\{u,v\}$ in the opposite direction):

\begin{equation}
   {  \mathcal{E}^t_{u,v}(c_i) \Rightarrow \bigvee_{j=1,2,...,k|j \neq i \wedge (u^t,v^{t+1}) \in E_j}{\mathcal{E}^t_{v,u}(c_j)}
   }
   \label{eq-swap}
\end{equation}
\vspace{0.25cm}

\noindent
{\bf TROT:} TROT is treated in a complementary way to TSWAP. Instead of adding constraints (\ref{eq-swap}) we add constraints forbidding simultaneous traversal in the opposite direction as follows:

\begin{equation}
   {  \mathcal{E}^t_{u,v}(c_i) \Rightarrow \bigwedge_{j=1,2,...,k|j \neq i \wedge (u^t,v^{t+1}) \in E_j}{\neg \mathcal{E}^t_{v,u}(c_j)}
   }
   \label{eq-rot}
\end{equation}
\vspace{0.25cm}

\section{COMBINING SAT-BASED APPROACH AND CBS}

Close look at CBS reveals that it operates similarly as problem solving in {\em satisfiability modulo theories} (SMT) \cite{DBLP:journals/constraints/BofillPSV12}. SMT divides satisfiability problem in some complex theory $T$ into an abstract propositional part that keeps the Boolean structure of the problem and simplified decision procedure $DECIDE_T$ that decides conjunctive formulae over $T$. A general $T$-formula is transformed to {\em propositional skeleton} by replacing atoms with propositional variables. The standard SAT-solving procedure then decides what variables should be assigned $TRUE$ in order to satisfy the skeleton - these variables tells what atoms holds in $T$. $DECIDE_T$ if the conjunction of satisfied atoms is satisfiable. If so then solution is returned. Otherwise conflict from $DECIDE_T$ is reported back and the skeleton is extended with a constraint forbidding the conflict.

The above observation let us to the idea to implement CBS in the SMT manner. The abstract propositional part working with the skeleton will be taken from MDD-SAT except that only constraints ensuring that assignments form valid paths interconnecting starting positions with goals will be preserved. Other constraints for collision avoidance will be omitted initially. Paths validation procedure will act as $DECIDE_T$ and will report back a set of conflicts found in the current solution. We call this algorithm SMT-CBS and it is shown in pseudo-code as Algorithm \ref{alg-SMTCBS} (it is formulated for MAPF; but is applicable for TSWAP, TPERM, and TROT after replacing conflict resolution part).

\begin{algorithm}[h]
\begin{footnotesize}
\SetKwBlock{NRICL}{SMT-CBS ($\Sigma = (G=(V,E),A,\alpha_0,\alpha_+))$}{end} \NRICL{
    $conflicts \gets \emptyset$\\
    $paths \gets$ $\{$shortest path from $\alpha_0(a_i)$ to $\alpha_+(a_i) | i = 1,2,...,k\}$ \\
    $\xi \gets \sum_{i=1}^k{\xi(paths(a_i))}$ \\
    \While {$True$}{
         $(paths,conflicts) \gets$ SMT-CBS-Fixed($conflicts,\xi,\Sigma$)\\
        \If {$paths \neq$ UNSAT}{
        	\Return $paths$\\
        }
        $\xi \gets \xi + 1$\\
    }
}   
 
\SetKwBlock{NRICL}{SMT-CBS-Fixed($conflicts,\xi,\Sigma$)}{end} \NRICL{
	    $\mathcal{F}(\xi)) \gets$ encode-Basic$(conflicts,\xi,\Sigma)$\\
	    \While {$True$}{
	        $assignment \gets$ consult-SAT-Solver$(\mathcal{F}(\xi))$\\
	        \If {$assignment \neq UNSAT$}{
	            $paths \gets$ extract-Solution$(assignment)$\\
	            $collisions \gets$ validate($paths$)\\
                   \If {$collisions = \emptyset$}{
                      \Return $(paths,conflicts)$\\
                   }
                   \For{each $(a_i,a_j,v,t) \in collisions$}{
                      $\mathcal{F}(\xi) \gets \neg \mathcal{X}_v^t(a_i) \vee \neg \mathcal{X}_v^t(a_j)$\\
                      $conflicts \gets conflicts \cup \{[(a_i,v,t),(a_j,v,t)]\}$
                   }
               }
               \Return {(UNSAT,$conflicts$)}\\
          }
}
\caption{Framework of SAT-based MAPF solving} \label{alg-SMTCBS}
\end{footnotesize}
\end{algorithm}

The algorithm is divided into two procedures: SMT-CBS representing the main loop and SMT-CBS-Fixed solving the input MAPF for a fixed cost $\xi$. The major difference from the standard CBS is that there is no branching at the high level. The high level SMT-CBS rougly correspond to the main loop of MDD-SAT. The set of conflicts is iteratively collected during entire execution of the algorithm. Procedure {\em encode} from MDD-SAT is replaced with {\em encode-Basic} that produces encoding that ignores specific movement rules (collisions between agents) but on the other hand encodes collected conflicts into $\mathcal{F}(\xi)$.

The conflict resolution in standard CBS implemented as high-level branching is here represented by refinement of $\mathcal{F}(\xi)$ with disjunction (line 20). Branching is thus deferred into the SAT solver. The presented SMT-CBS process builds in fact equisatisfiable formula to that built by MDD-SAT. The advantage of SMT-CBS is that it builds the formula lazily; that is, it adds constraints on demand after conflict occurs. Such approach may save resources as solution may be found before all constraint are added.

\section{EXPERIMENTAL EVALUATION}

We performed an extensive evaluation of all presented algorithms on standard synthetic benchmarks \cite{DBLP:conf/ijcai/BoyarskiFSSTBS15,DBLP:journals/ai/SharonSGF13}. Representative part of results is presented in this section.

\subsection{Benchmarks and Setup}

We implemented SMT-CBS in C++ on top of the Glucose 4 SAT solver \cite{DBLP:conf/sat/AudemardLS13,DBLP:conf/ijcai/AudemardS09} that ranks among the best SAT solvers according to recent SAT solver competitions \cite{DBLP:conf/aaai/BalyoHJ17}. The standard CBS has been re-implemented from scratch since the original implementation written in Java does support only grids but not general graphs \cite{SharonSFS15}. To obtain MDD-SAT applicable on TSWAP, TPERM, and TROT we modified the existing C++ implementation \cite{SurynekFSB16}. All experiments were run on an i7 CPU 2.6 Ghz under Kubuntu linux 16 with 8GB RAM. \footnote{To enable reproducibility of presented results we provide complete source code of our solvers on author's web: \texttt{http://users.fit.cvut.cz/$\sim$surynpav}.}

\begin{figure}[h]
    \centering
    \includegraphics[trim={4cm 25cm 11cm 3.3cm},clip,width=0.4\textwidth]{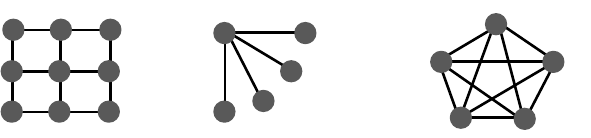}
    \vspace{-0.2cm}\caption{Example of regular 4-connected {\em grid}, {\em star}, and {\em clique}.}
    \label{figure-types}
\end{figure}

The experimental evaluation has been done on diverse instances consisting of 4-connected {\em grid} of size $8 \times 8$ and $16 \times 16$, {\em random graphs} containing 20\% of random edges, {\em star} graphs, and {\em cliques} (see Figure \ref{figure-types}). Initial and goal configurations of tokens/agents have been generated randomly in all tests.

\begin{figure}[h]
    \centering
    \includegraphics[trim={2.5cm 14.5cm 2cm 2.5cm},clip,width=0.5\textwidth]{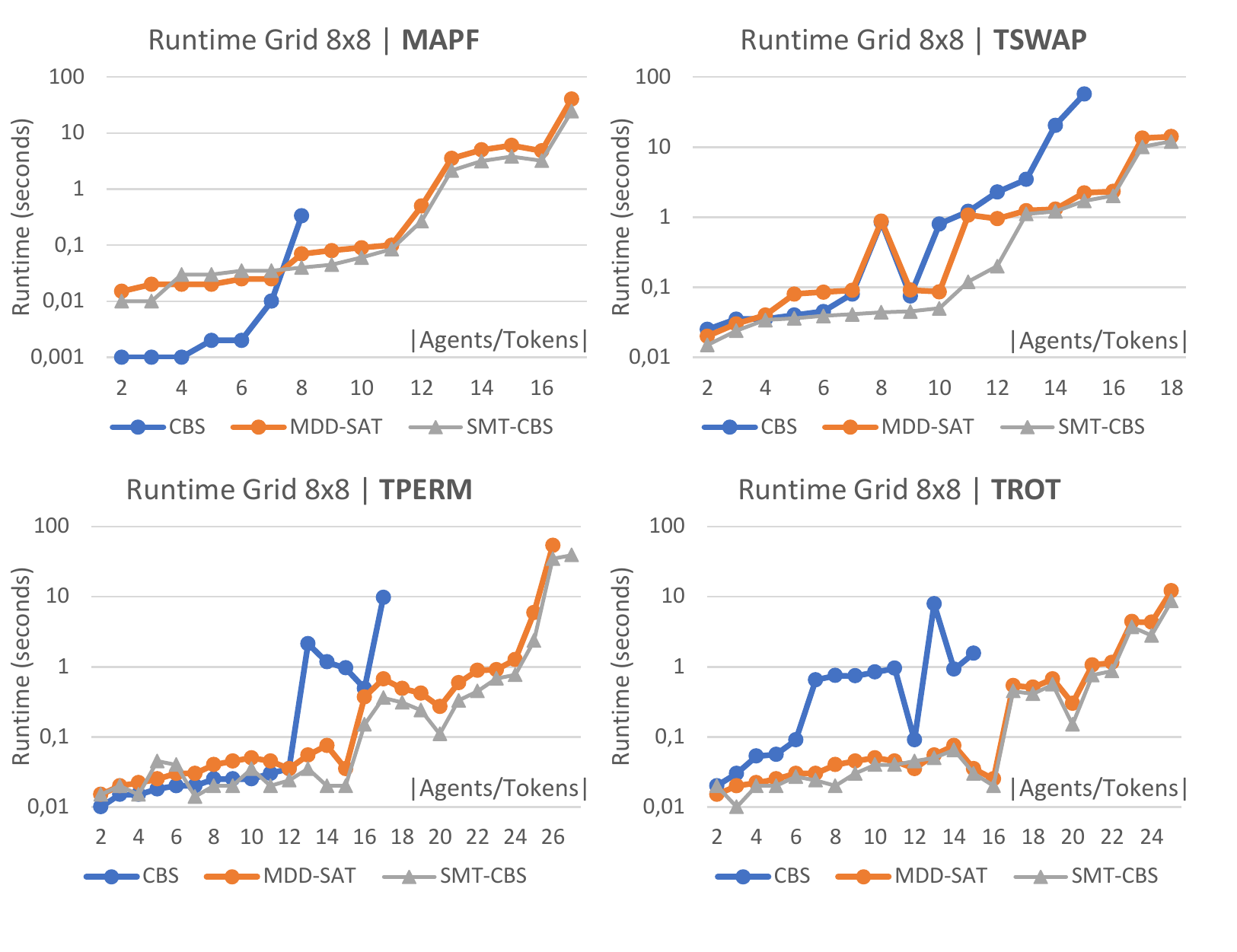}
    \vspace{-1.0cm}\caption{Runtime comparison of CBS, MDD-SAT, and SMT-CBS algorithms solving MAPF, TSWAP, TPERM, and TROT on $8 \times 8$ {\em grid}.}
    \label{exp-grid-8x8}
\end{figure}

We varied the number of items in relocation instances to obtain instances of various difficulties; that is, the underlying graph was not fully occupied - which in MAPF has natural meaning while in token problems we use one special color $\bot \in C$ that stands for an empty vertex (that is, we understand $v$ as empty if and only if $\tau(v)=\bot$). For each number of items we generated 10 random instances - the mean value out of these 10 instances is always presented. The timeout in all test was set to 60 seconds. Presented results were obtained from instances solved within this timeout.

\subsection{Comparison of Algorithms}

\begin{figure}[h]
    \centering
    \includegraphics[trim={2.6cm 21cm 2.5cm 2.5cm},clip,width=0.5\textwidth]{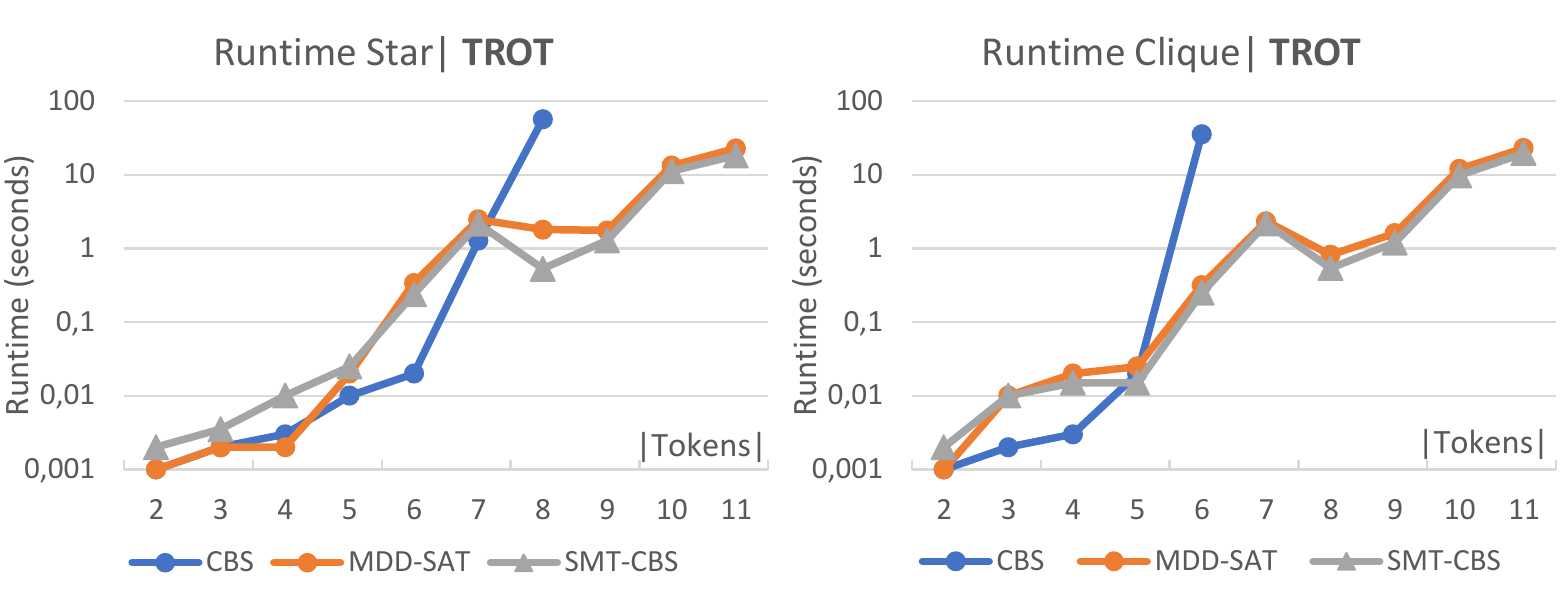}
    \vspace{-0.6cm}\caption{Comparison of TROT solving by CBS, MDD-SAT, and SMT-CBS on a {\em star} and {\em clique} graphs consisting of 16 vertices.}
    \label{exp-rot}
\end{figure}

\begin{figure}[h]
    \centering
    \includegraphics[trim={2.4cm 24.2cm 6cm 2.5cm},clip,width=0.5\textwidth]{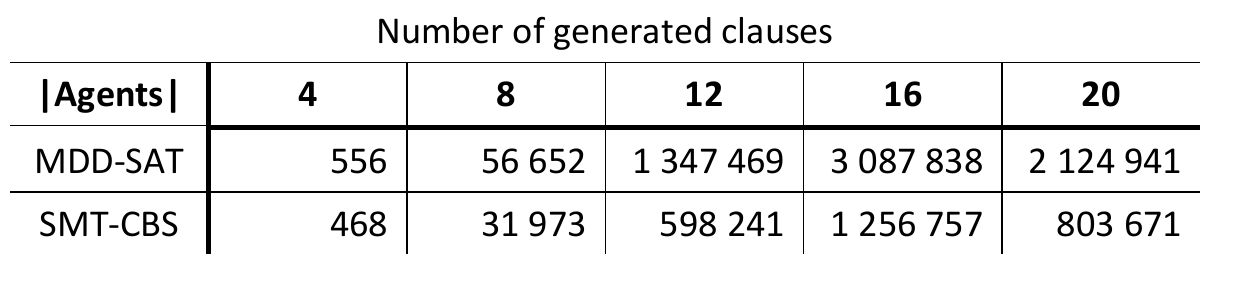}
    \vspace{-0.6cm}\caption{Comparison of the size of encodings generated by MDD-SAT and SMT-CBS (number of clauses is shown) on MAPF instances.}
    \label{exp-clauses}
\end{figure}

Our tests were focused on the runtime comparison and evaluation of the size of encodings in case of MDD-SAT and SMT-CBS. Part of results we obtained is presented in Figures \ref{exp-grid-8x8}, \ref{exp-rot}, and \ref{exp-clauses}. In all tests CBS turned out to be incompetitive against MDD-SAT and SMT-CBS on instances containing more agents. This is expectable result as it is known that performance of CBS degrades on densely occupied instances \cite{surynek2016empirical}.

SMT-CBS turned out to be fastest in performed tests. It reduces the runtime by about $30\%$ to $50\%$ relatively to MDD-SAT. More significant benefit of SMT-CBS was observed in MAPF and TSWAP while in TROT and TPERM the improvement was less significant. Both MAPF and TSWAP have more clauses in their encodings used by MDD-SAT than TROT and TPERM on the instance hence SMT-CBS has greater room for reducing the size of encoding by constructing it lazily in these types of relocation problems. This claim has been experimentally verified (Figure \ref{exp-clauses}); the SMT-CBS reduces the number of clauses to less than half of the number generated by MDD-SAT.

\section{CONCLUSIONS}

We introduced a general framework for reasoning about item relocation problems in graphs based on concepts from the CBS algorithm. In addition to two known problems MAPF and TSWAP we introduced two derived variants TROT and TPERM in this context. We also suggested novel algorithm SMT-CBS inspired by satisfiability modulo theories that combines SAT-based approach with CBS. Experimental evaluation showed that SMT-CBS significantly outperforms previous state-of-the-are SAT-based solver MDD-SAT.

Experimental evaluation indicates that SMT-CBS significantly outperforms MDD-SAT, previous state-of-the-art SAT-based solver for MAPF, on all studied relocation problems. The most significant benefit of SMT-CBS can be observed on higly constrained MAPF and TSWAP instances. For future work we plan to further reduce the size of SAT encodings in SMT-CBS by eliminating unnecessary time expansions in MDDs.

\addtolength{\textheight}{-1cm}   % This command serves to balance the column lengths
                                  % on the last page of the document manually. It shortens
                                  % the textheight of the last page by a suitable amount.
                                  % This command does not take effect until the next page
                                  % so it should come on the page before the last. Make
                                  % sure that you do not shorten the textheight too much.

%%%%%%%%%%%%%%%%%%%%%%%%%%%%%%%%%%%%%%%%%%%%%%%%%%%%%%%%%%%%%%%%%%%%%%%%%%%%%%%%

%%%%%%%%%%%%%%%%%%%%%%%%%%%%%%%%%%%%%%%%%%%%%%%%%%%%%%%%%%%%%%%%%%%%%%%%%%%%%%%%

%%%%%%%%%%%%%%%%%%%%%%%%%%%%%%%%%%%%%%%%%%%%%%%%%%%%%%%%%%%%%%%%%%%%%%%%%%%%%%%%
%\section*{APPENDIX}
%Appendixes should appear before the acknowledgment.

%\section*{ACKNOWLEDGMENT}
%This work is supported by Faculty of Information Technology, Czech Technical University in Prague.

%%%%%%%%%%%%%%%%%%%%%%%%%%%%%%%%%%%%%%%%%%%%%%%%%%%%%%%%%%%%%%%%%%%%%%%%%%%%%%%%

\bibliographystyle{IEEEtran}
\bibliography{bibfile}

\vfill

\end{document}